\pdfoutput=1

\documentclass[11pt]{article}

\usepackage[]{acl}

\usepackage{times}
\usepackage{latexsym}
\usepackage{bm}

\usepackage[T1]{fontenc}

\usepackage[utf8]{inputenc}

\usepackage{microtype}
\usepackage{amsthm,amsmath,amssymb}
\usepackage{mathrsfs}
\usepackage{booktabs}
\usepackage{multirow}
\usepackage{mathrsfs}
\usepackage{colortbl}
\usepackage{graphicx}  
\usepackage{float}  
\usepackage{subfigure} 
\usepackage{soul}
\definecolor{mygray}{gray}{.95}

\title{Semantic-based Pre-training for Dialogue Understanding}
\author{
 Xuefeng Bai\thanks{~~Work done as an intern at Tencent AI Lab.}$^{*\spadesuit \diamondsuit}$\hspace{0.5mm}, 
 Linfeng Song$^{\clubsuit}$\hspace{0.5mm}, 
 Yue Zhang$^{\spadesuit \diamondsuit}$\hspace{0.2mm}\hspace{1.5mm} \\
 $^\spadesuit$ School of Engineering, Westlake University, China\\
 $^\clubsuit$ Tencent AI Lab, Bellevue, WA, USA \\
 $^\diamondsuit$ Institute of Advanced Technology, Westlake Institute for Advanced Study, China 
}

\begin{document}
\maketitle
\begin{abstract}
Pre-trained language models have made great progress on dialogue tasks.
However, these models are typically trained on surface dialogue text, thus are proven to be weak in understanding the main semantic meaning of a dialogue context.
We investigate Abstract Meaning Representation (AMR) as explicit semantic knowledge for pre-training models to capture the core semantic information in dialogues during pre-training.
In particular, we propose a semantic-based pre-training framework that extends the standard pre-training framework \cite{devlin-etal-2019-bert} by three tasks for learning 1) core semantic units, 2) semantic relations and 3) the overall semantic representation according to AMR graphs.
Experiments on the understanding of both chit-chats and task-oriented dialogues show the superiority of our model.
To our knowledge, we are the first to leverage a deep semantic representation for dialogue pre-training.

\end{abstract}

\section{Introduction}
Dialogue systems have attracted increasing attention from both academia and industry researches~\cite{ChenLYT17,deriu2021survey,GAO2021100}.
The tasks can be commonly divided into two categories: task-oriented dialogue systems~\cite{wen-etal-2017-network,dinan2018wizard,mehri2020dialoglue} and chit-chat dialogue systems~\cite{ritter-etal-2011-data,li-etal-2017-dailydialog,yu-2020-dialogue,cui-etal-2020-mutual,chen-etal-2021-dialogsum,chen-etal-2022-Crosslingual,Song22CASA}. 
The former aims to interact in the context of a specific task, while the latter chats with users without task and domain restrictions.
Despite differences in goals, a common challenge for both tasks is understanding the semantic information conveyed in a dialogue history.

Recently, semantic representations from pre-trained language models have achieved remarkable success on a spectrum of dialogue tasks~\cite{wen2015semantically,zhang-etal-2020-dialogpt,wu-etal-2020-tod,GuTLXGJ20,ZengYLGS21,zhang-zhao-2021-structural,cui-etal-2021-knowledge}, where knowledge learned in pre-training over large-scale dialogue corpora can be transferred to downstream applications.
Current pre-training techniques typically focus on the surface text.
However, they do not explicitly consider deep semantic clues beyond text,
which leads to some unexpected behavior, such as paying attention to meaningless words~\cite{mudrakarta-etal-2018-model}, and suffering from spurious feature associations~\cite{Kaushik2020Learning} and adversarial attacks~\cite{JiaL17Adv}.

Incorporating semantic information into dialogue systems has been shown to be helpful for many downstream tasks, such as dialogue intent prediction~\cite{gupta-etal-2018-semantic-parsing}, dialogue state tracking~\cite{ChengAABDFKKLPW20}, and dialogue relation extraction~\cite{bai-etal-2021-semantic}.
These methods first parse dialogue turns into semantic structures, and then incorporate them as extra features into neural systems.
However, they 1) only focus on domain-specific benchmark data, leaving the general potentiality of semantic structures unexploited;
2) require either human annotations or an external parser to obtain semantic structures, raising costs or/and causing error propagation for real applications.

\begin{figure}
    \centering
    \includegraphics[width=0.35\textwidth]{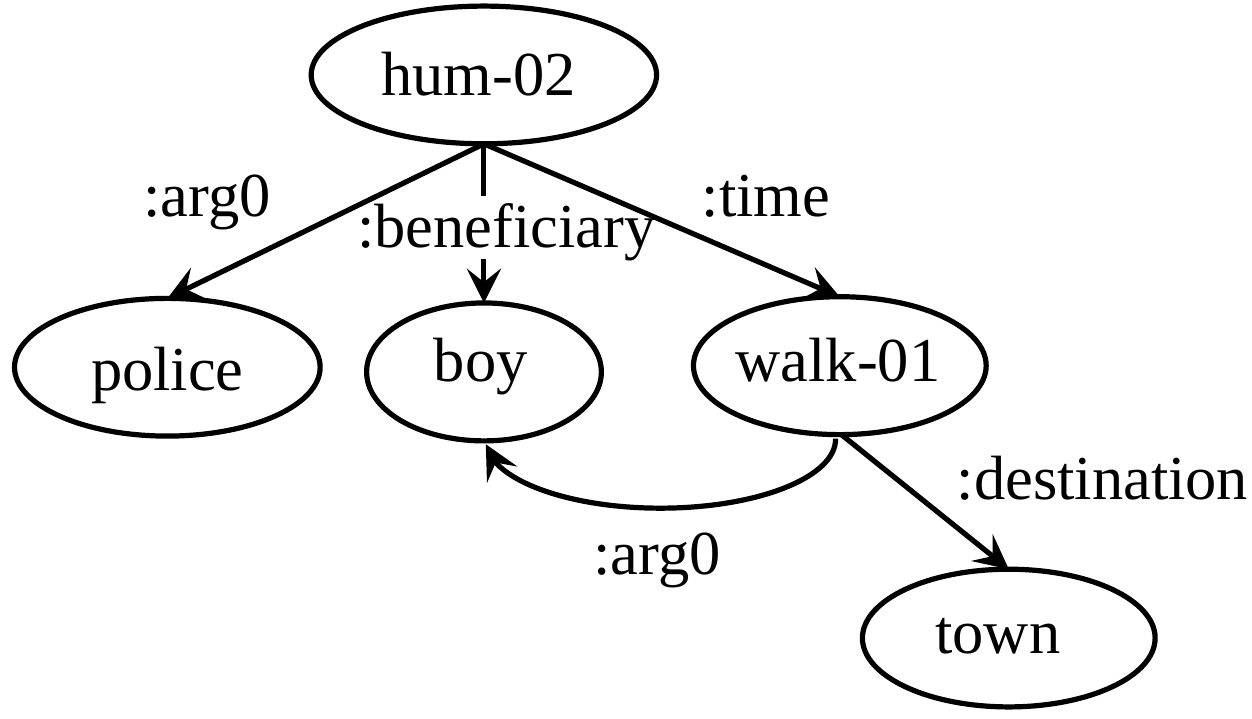}
    \caption{An AMR graph for sentence ``\textit{The police hummed to the boy as he walked to town.}''}
    \label{fig:intro-example}
\end{figure}

We present \textbf{SARA}, a \textbf{S}emantic-graph-based pre-tr\textbf{A}ining f\textbf{R}amework for di\textbf{A}logues, aiming to endow a pre-trained dialogue model with a stronger ability to infer semantic structures from conversations
by using explicit semantic structures for more fine-grained supervisions.
In particular, we exploit the abstract meaning representation (AMR;~\citeauthor{banarescu2013abstract}~\citeyear{banarescu2013abstract}), a fine-grained deep structure widely adopted in semantic parsing~\cite{lyu-titov-2018-amr,zhang-etal-2019-broad,cai-lam-2020-amr,Bevilacqua_Blloshmi_Navigli_2021,bai-etal-2022-graph} and generation~\cite{konstas2017neural,song2018graph,zhu2019modeling,bai-etal-2020-online,ribeiro-etal-2021-structural}.
As shown in Figure~\ref{fig:intro-example}, AMR represents a sentence using a rooted directed graph, highlighting the core semantic units (e.g., ``\textit{police}'', ``\textit{hum}'', ``\textit{boy}'') in a sentence and connecting them with semantic relations (e.g., ``\textit{:arg0}'', ``\textit{:time}''). 

\begin{figure*}
    \centering
    \begin{minipage}[b]{0.25\textwidth}
    \subfigure[Example AMR graph.]{
        \includegraphics[width=1\textwidth]{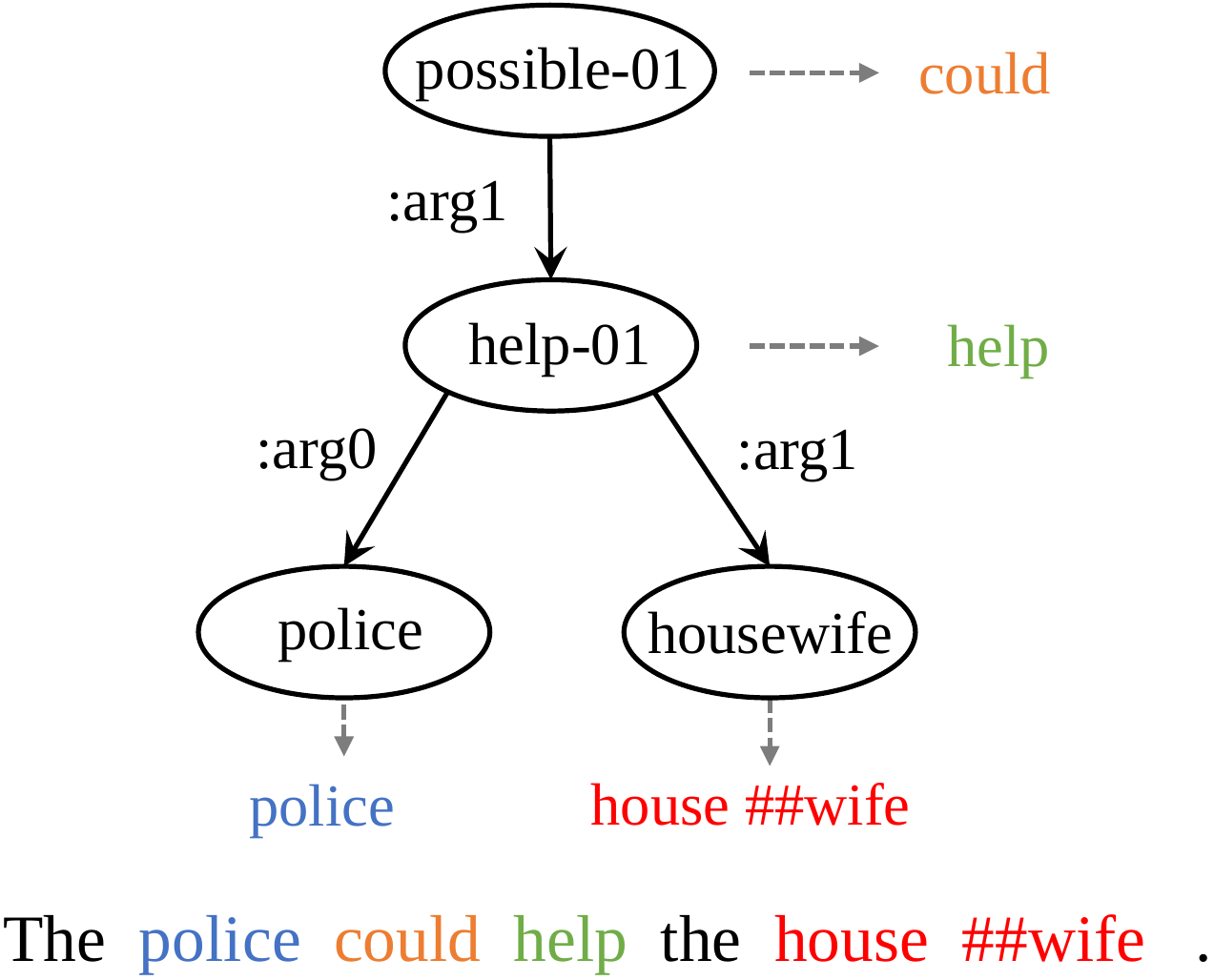}
        \label{fig:example}
    }
    \end{minipage}
    \hspace{2mm}
    \begin{minipage}[b]{0.35\textwidth}
    \centering
    \subfigure[Semantics-guided masking]{
        \includegraphics[width=1.05\textwidth]{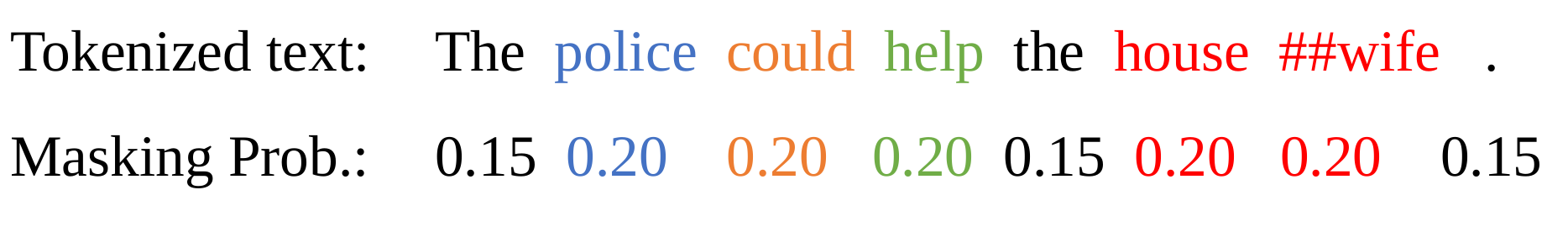}
        \label{fig:taskmlm}
    }
    \subfigure[Semantic relation prediction]{
        \includegraphics[width=1.1\textwidth]{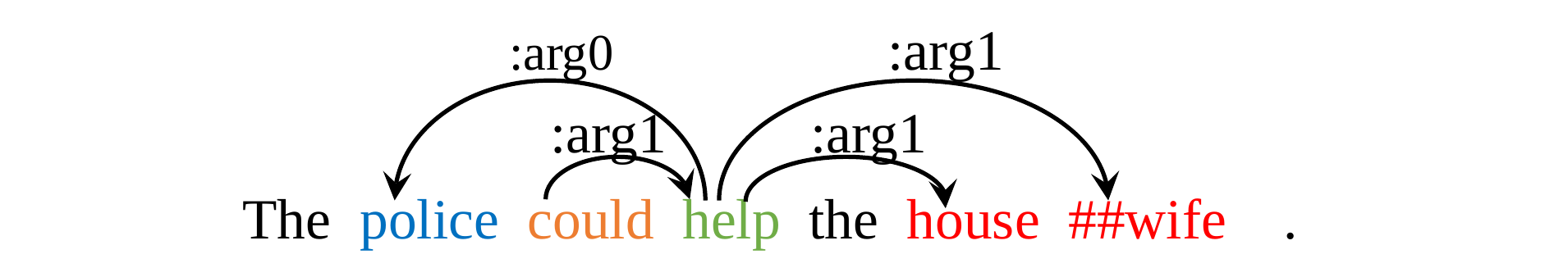}
        \label{fig:taskrel}
    }
    \end{minipage}
    \hspace{2mm}
    \begin{minipage}[b]{0.34\textwidth}
    \subfigure[Semantic agreement]{
        \includegraphics[width=1\textwidth]{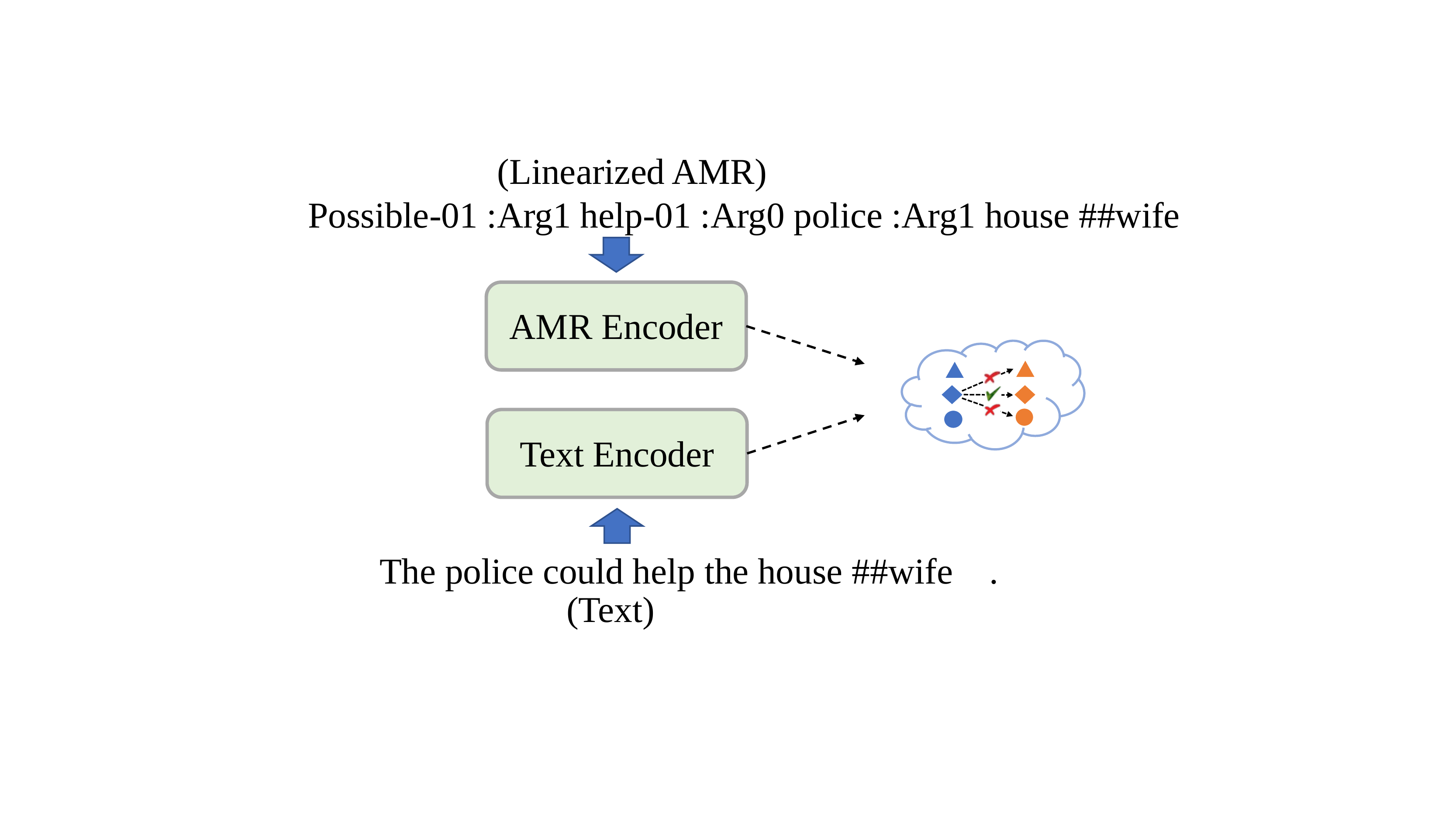}
        \label{fig:tasksim}
    }
    \end{minipage}
    \vspace{-3mm}
    \caption{The semantic-based pre-training framework.}
    \label{fig:model}
\end{figure*}

We explicitly leverage AMR graphs for pre-training our dialogue model.
As shown in Figure~\ref{fig:model}, {SARA} consists of three pre-training sub-tasks: 1) semantic-based mask language modeling, which extends the standard mask language modeling task~\cite{devlin-etal-2019-bert} by paying more attention to core semantic units in a dialogue; 2) semantic relation prediction, which aims to learn semantic relations between words; 3) semantic agreement, which optimizes the overall similarity between a dialogue and its corresponding AMR graph.
The {SARA} combines strengths of both powerful contextualized representation of pre-trained models and explicit semantic knowledge, while eliminating the requirement of an external semantic parser in downstream applications.

We choose BERT~\cite{devlin-etal-2019-bert} and \textsc{RoBERTa}~\cite{Liu19Roberta} models as backbone, which are then continual pre-trained on a large-scale conversation dataset using our framework.
Experiments show that our semantic-based framework gives better results than current pre-training methods that use much more training data, achieving new state-of-the-art results on both chit-chat understanding (dialogue relation extraction) and task-oriented dialogue understanding tasks (DialoGLUE benchmark).
Our method also gives better results than previous semantic-base systems on downstream tasks, without using an external parser.
Further analysis suggests that semantic information introduced by AMR can help our model to better understand semantically complex dialogues.
To our knowledge, we are the first to leverage deep semantic representation for dialogue pre-training. 
Our code and the pre-trained models are available at \url{https://github.com/goodbai-nlp/Sem-PLM}. 

\section{Related Work}
\noindent\textbf{Pre-training for Dialogue.}
Inspired by the success of pre-trained language models in the general domain~\cite{peters-etal-2018-deep, Radford2018ImprovingLU,devlin-etal-2019-bert,lewis-etal-2020-bart}, various pre-trained models have been proposed in the domain of dialogue. 
DialoGPT~\cite{zhang-etal-2020-dialogpt} continual pre-trains a GPT-2~\cite{Radford2019LanguageMA} model directly on Reddit comments data.
ConvRT~\cite{henderson-etal-2019-repository} pre-trains a dual Transformer encoder for the response selection task.
PLATO~\cite{bao-etal-2020-plato} introduces a latent variable-based model for dialogue response generation pre-training.
TOD-BERT~\cite{wu-etal-2020-tod} pre-trains a Transformer encoder on task-oriented dialogue corpus for task-oriented dialogue applications.
MPC-BERT~\cite{GuTLXGJ20} continues to pre-train a BERT model with self-supervised tasks based on the interactions among utterances and interlocutors.
SPIDER~\cite{zhang-zhao-2021-structural} continues to pre-train a BERT model with auxiliary tasks to predict the utterance order and understand the sentence backbone.
DialogLM~\cite{Zhong22DialogLM} pre-trains a generative Transformer encoder on long conversations with window-based pre-training tasks.
Our work is similar in that we also pre-train a model on the dialogue corpora.
However, unlike these previous studies, which focus on text level distributions, we additionally enhance the model with semantic structures.

\noindent\textbf{Semantics for dialogue.}
Semantic knowledge has been used for both social chat and task-oriented dialogues systems. 
PEGASUS~\cite{zue-etal-1994-pegasus} transforms a sentence into a semantic frame which is then used for travel planing.
\citet{WirschingHuberKoelbletal2012} design a dialogue system which performs database operations based on semantic features.
\citet{gupta-etal-2018-semantic-parsing} and \citet{aghajanyan-etal-2020-conversational} integrate intents and slots into a semantic tree and solve intent classification and slot-filling tasks as semantic parsing.
\citet{ChengAABDFKKLPW20} represent task-oriented dialogue as a semantic graph to perform dialogue state tracking.
A most related work is~\citet{bai-etal-2021-semantic}, who build dialogue-level AMR graphs for both social chat understanding and dialogue response generation.
Our work is similar in showing the effect of semantic knowledge for improving dialogue understanding.
However, different from them, we focus on enhancing the language model with semantic knowledge during pre-training, and our model does not require an external AMR parser in downstream applications.


\section{Method}

Figure~\ref{fig:model} illustrates our semantic-based pre-training framework for dialogues.
We take a pre-trained Transformer~\cite{vaswani2017attention} encoder as the backbone, using AMR as explicit semantic knowledge to continuously pre-train the model on dialogues in a multitask setting.
In particular, the following three semantic-aware tasks are designed: \\
\indent\textbullet~Semantics-based masking (Section~\ref{sec:sem-mlm}). \\
\indent\textbullet~Semantic relation prediction (Section~\ref{sec:rel-pred}). \\
\indent\textbullet~Semantic agreement (Section~\ref{sec:overallsim}).

The former two learn semantic knowledge from AMR nodes and AMR edges, respectively.
The last task regularizes the overall representation of a dialogue using graph-level semantic features.

We follow~\citet{bai-etal-2021-semantic} and construct dialogue-level AMR graphs by $2$ steps: 1) building utterance-level AMR graphs by independently transforming utterances into AMR using a pre-trained AMR parser.
2) connecting utterance-level AMR graphs with a root node, where
edges are labeled with the corresponding speaker.

Formally, denote an input dialogue sequence\footnote{Please refer to Appendix~\ref{sec:appendix-b} for dialogue input format.} as $\bm{x} = [x_1, x_2, ..., x_n]$, 
where $n$ is the number of tokens in the dialogue. The corresponding AMR is a directed acyclic graph $\mathcal{G} = \left\langle\mathcal{V}, \mathcal{E}\right\rangle$, where $\mathcal{V}$ denotes a set of nodes (i.e., AMR concepts) and $\mathcal{E}$ (i.e., AMR relations) denotes a set of labeled edges.
An edge can be further represented by a triple $\left\langle v_i, r_{ij}, v_j\right\rangle$, meaning that the edge is from node $v_i$ to $v_j$ with label $r_{ij}$.

\subsection{Task 1: Semantics-guided Masking}
\label{sec:sem-mlm}
We first formally present the vanilla mask language modeling (MLM) setup, before introducing the semantic-guided masking strategy.

\noindent\textbf{Vanilla MLM}. Given a sequence of tokens $\bm{x}$, the standard masking strategy~\cite{devlin-etal-2019-bert} selects a set fraction of tokens positions (denoted as $\bm{m}=[m_1,m_2,...,m_k]$) for masking independently at random, and use these ``\textit{selected}'' tokens $\{x_i| i \in \bm{m} \}$ as supervisions to train a language model. Formally, denoting the masked text as $\bm{\tilde{x}}$, vanilla MLM optimizes the following training objective:
\begin{equation}
    \ell_{vanilla\_mlm} = -\sum_{i \in \bm{m}} \text{log} P(x_i|\bm{\tilde{x}}),
    \label{eq:mlm}
\end{equation}
where the conditional probability $P(x_i|\bm{\tilde{x}})$ is generated by an encoder model with a softmax layer.

\noindent\textbf{Semantics-guided Masking}. A salient limitation of vanilla MLM is that it treats all tokens equally, thus potentially wasting resources on tokens that provide little signal (e.g., punctuations and stop words). 
We introduce a semantic-guided masking strategy, encouraging model to give more attention on \textit{semantic-aware} units, which are expected to have more influence on text understanding.
As shown in Figure~\ref{fig:taskmlm}, our semantic-guided masking strategy gives a higher masking probability for tokens (e.g. ``\textit{police}'', ``\textit{could}'', ``\textit{help}'') that contain important semantic information.
Formally, we define a token as a \textit{semantic-aware} unit when it is aligned with an AMR node, according to the AMR-to-text alignment $\mathcal{A}$\footnote{$\mathcal{A}$ is a one-to-$K$ mapping ($K\in [1,\dots, n]$).} (An example is given in Figure~\ref{fig:example}).
Since pre-trained models typically use a vocabulary with sub-word units~\cite{sennrich-etal-2016-neural}, for an alignment pair $\left \langle v_i, x_j  \right \rangle$, we extend the alignment as  $\left \langle v_i, \{x_j^1, x_j^2...,x_j^l\}  \right \rangle$, where the AMR node $v_i$ is aligned a set of all tokens $\{x_j^1 ,x_j^2...,x_j^l\}$ which are sub-words of word $w_j$.
For example, in Figure~\ref{fig:model}, the AMR node ``\textit{housewife}'' is aligned with sub-tokens ``\textit{house}'' and ``\textit{\#\#wife}''.

Denoting $\bm{m'}=[m'_1,m'_2,...,m'_k]$ as token indices selected by the proposed semantic-guided masking strategy, the training objective is:
\begin{equation}
    \ell_{sem\_mlm} = -\sum_{i \in \bm{m'}} \text{log} P(x_i|\bm{\tilde{x}}).
    \label{eq:sem-mlm}
\end{equation}

We follow \textsc{RoBERTa}~\cite{Liu19Roberta} and use the dynamic masking, where we generate the masking pattern every step instead of performing masking during data preprocessing.

\subsection{Task 2: Semantic Relation Prediction}
\label{sec:rel-pred}
The semantic relation prediction task is designed for learning the semantic relations between words. 
To this end, we project the edges of each input AMR graph onto the
corresponding sentence according to their node-to-word alignments (as shown in Figure~\ref{fig:taskrel}), before training a predictor to generate the projected edges.

\noindent\textbf{Relation Projection}. 
Since AMR relations are defined on AMR nodes instead of words in the dialogue text, we use a node-to-word alignment $\mathcal{A}$ to project the AMR edges $\mathcal{E}$ onto text with following rules:
\begin{equation}
	\hat{r}_{ij}=\left\{
	\begin{aligned}
		r_{i'j'} &, &\text{if}~x_{i} \in \mathcal{A}(v_{i'}), x_{j} \in \mathcal{A}(v_{j'}), \\
		\textit{None} &,& \text{otherwise} \text{.}
	\end{aligned}
	\right.
	\label{eq:projectededge}
\end{equation}

The same strategy in Section~\ref{sec:sem-mlm} is used to deal with sub-word tokens.

\noindent\textbf{Relation Prediction}.
We first use a Transformer encoder to generate contextualized word hidden states $\bm{h}=[h_1,h_2,...,h_n]$.
Based on that, a deep biaffine neural parser~\cite{dozat2016deep} is used to predict the relations between words. 
To determine whether a directed edge (or \textit{arc}) from $x_i$ to $x_j$ exists, the biaffine parser first uses two separate MLPs (denoted as $\texttt{MLP}^H$ and $\texttt{MLP}^D$) to obtain two lower-dimensional representation vectors for each position, then calculates scores via a biaffine operation:
\begin{equation}
    \begin{aligned}
    r_i^H, r_j^D &= \texttt{MLP}^H(h_i), \texttt{MLP}^D(h_j), \\
    \textit{s}_{ij}^{arc} &= {\begin{bmatrix} r_j^D \\ 1 \end{bmatrix}}^T W^{arc}  r_i^H, \\
    P(y_{ij}^{arc}|\bm{x}) &= \text{softmax}_j(\textit{s}_{i}^{arc}),
    \end{aligned}
    \label{eq:biaffine}
\end{equation}
where $r_i^H$ is the representation vector of $x_i$ as a head word, and $r_j^D$ denotes the vector of $x_j$ as a dependent word. $P(y_{ij}^{arc}|\bm{x})$ is the probability of the \textit{arc} $(i,j)$, and $W^{arc}$ is a parameter matrix.
To calculate the probability of assigning a label $l$ to the \textit{arc}$(i,j)$, which is denoted as $P(y_{ijl}^{label}|\bm{x})$, the biaffine parser uses the same scorer as in Equation~\ref{eq:biaffine} but with different parameters for MLPs and biaffines.\footnote{The biaffine parameter for label scoring is a three dimensional tensor.}

The training objective of relation prediction is:
\begin{equation}
    \ell_{rel} = - \sum_{\left \langle x_i, \hat{r}_{ij}, x_j  \right \rangle \in \mathcal{E}'}\text{log}~P(y_{ij}^{arc}|\bm{x})P(y_{ij\hat{r}_{ij}}^{label}|\bm{x}),
    \label{eq:rel-pred}
\end{equation}
where $\mathcal{E}'$ represents the \textit{projected} AMR edges. 

\begin{table*}[!t]
	\centering
	\small
	\begin{tabular}{lcccccccc}
		\toprule
		\textbf{Dataset} & \textbf{DialogRE} & \textbf{\textsc{banking77}} &\textbf{\textsc{hwu64}} &\textbf{\textsc{clinc150}} &\textbf{\textsc{rest8k}} & \textbf{\textsc{DSTC8}} & \textbf{\textsc{TOP}} & \textbf{\textsc{MultiWOZ}} \\
		\midrule
		train &5,997 &8,622 &8,954 &15,000 &7,244 &5,023 &31,279 &56,774 \\
		dev &1,914 &1,540 &1,076 &3,000 &1,000 &602 &4,462 &7,374 \\
		test &1,862 &3,080 &1,076 &4,500 &3,731 &1,813 &9,042 &7,372 \\
		\bottomrule
	\end{tabular}
	\caption{Statistics of datasets.}
	\label{tab:statistics}
\end{table*}

\subsection{Task 3: Semantic Agreement}
\label{sec:overallsim}
We encourage the model to learn the overall agreement of a dialogue and its corresponding AMR graph.
As shown in Figure~\ref{fig:tasksim}, we use an auxiliary network to encode the AMR, and maximize the similarity score between the hidden states of text and AMR.
Following previous work~\cite{konstas2017neural}, we linearize AMR graphs into a sequence (refer to Figure~\ref{fig:tasksim} for an example) and use a pre-trained encoder to transform AMR into a set of hidden states.\footnote{We also tried a structure-aware encoder but without observing significant improvements.}

Formally, defining the linearized AMR graph as $\bm{g} = [g_1, g_2,...,g_m]$, the vector representation of text and its corresponding AMR is calculated as:
\begin{equation}
    \begin{aligned}
        h^{\text{text}} &= \texttt{Pooling}(\texttt{TextEnc}(\bm{x})), \\
        h^{\text{amr}} &= \texttt{Pooling}(\texttt{TextEnc}(\bm{g})), \\
    \end{aligned}
    \label{eq:biencoder}
\end{equation}
where $\texttt{TextEnc}(\cdot)$ and $\texttt{TextEnc}(\cdot)$ are text encoder and AMR encoder, respectively. They are initialized with the same weights but updated separately during training.  $\texttt{Pooling}(\cdot)$ is a function that reduces that sequence of vectors into one vector. Following BERT~\cite{devlin-etal-2019-bert}, we feed the hidden state of the first input token into a MLP layer to get the ``pooled'' vector.

We use the cosine similarity as a distance scoring function and adopt the contrastive learning framework~\cite{Hadsell06dim,FrosstPH19,GaoYC21,luo-etal-2022-Mere} to train our model, with the aim to pulling semantically close text-AMR pairs and pushing apart unpaired examples. In particular, for a given text $\bm{x}$, the \textit{positive} example is its corresponding AMR graph $\bm{g}$, the \textit{negative} examples are the AMR graphs of its neighbor dialogues in the corpus.
Formally, let $h^{\text{text}}_i$ and $h^{\text{amr}}_i$ denote the representations of the $i$th $\left\langle\textit{text},\textit{AMR}\right\rangle$ pair in the dataset, the training objective is:
\begin{equation}
    \ell_{sim} = - \text{log} \frac{\text{exp}(\text{sim}(h^{\text{text}}_i, h^{\text{amr}}_i)/\tau)}{\sum_{j\in \mathcal{N}(i)} \text{exp}(\text{sim}(h^{\text{text}}_i, h^{\text{amr}}_j)/\tau)},
    \label{eq:contrastsim}
\end{equation}
where $\text{sim}(\cdot,\cdot)$ denotes the cosine similarity, $\mathcal{N}(i)$ collects neighbor index of the $i$th example, and $\tau > 0$ denotes the temperature hyper-parameter.

\subsection{Training}
\label{sec:training}
Our model is trained by optimizing the total loss of above $3$ tasks:
\begin{equation}
    \ell_{total} = \ell_{sem\_mlm} + \alpha\ell_{rel} + \beta\ell_{sim},
    \label{eq:total-loss}
\end{equation}
where $\alpha$ and $\beta$ are weighting hyper-parameters for $\ell_{rel}$ and $\ell_{sim}$, respectively.
To make the computational requirements feasible, we do not train our model from scratch, but rather continue training a model that has been pre-trained on textual inputs.
Our framework is architecture-flexible and can be be applied to different models such as BERT, \textsc{RoBERTa}, and BART.

\section{Experiments}
We evaluate the effectiveness of our semantic pre-training model on $8$ dialogue tasks and compare the results with the state-of-the-art pre-trained and semantic-enriched models.

\subsection{Dataset}
\noindent\textbf{Pre-training Corpus.} We continual pre-train our model on the Reddit~\cite{henderson-etal-2019-repository} corpus. 
After sampling and filtering (refer Appendix~\ref{sec:appendix-a}), 
the dataset comprises 5,864,254 dialogue instances, in total 397 million words.
We adopt the state-of-the-art AMRBART~\cite{bai-etal-2022-graph} parser \footnote{\url{https://github.com/muyeby/AMRBART}} to transform the text into AMR graphs. 
To obtain the AMR-to-text alignment, we use the JAMR aligner\footnote{\url{https://github.com/jflanigan/jamr}} released by~\citet{flanigan-etal-2014-discriminative}.

\noindent\textbf{Dialogue task datasets.} We evaluate our model on both chitchat and task-oriented understanding tasks. 
For chitchat, we focus on the dialogue relation extraction task which aims to predict the relationship between an given entity pair. 
We report results on both original (v1) and updated (v2) English version of \textbf{DialogRE}~\cite{yu-2020-dialogue}.

For task-oriented dialogue,  we report results on the DialoGLUE~\cite{mehri2020dialoglue} benchmark, which consists of 7 different datasets spanning 4 different tasks:  1) intention prediction, including \textbf{\textsc{banking77}}~\cite{casanueva-etal-2020-efficient}, \textbf{\textsc{clinc150}}~\cite{larson-etal-2019-evaluation} and \textbf{\textsc{hwu64}}~\cite{liu2021benchmarking}; 2) slot filling, including \textbf{\textsc{restaurant8k}}~\cite{coope-etal-2020-span} and \textbf{\textsc{DSTC8}}~\cite{rastogi-etal-2019-scaling}; 3) semantic parsing, \textbf{\textsc{TOP}}~\cite{gupta-etal-2018-semantic-parsing}; and 4) dialogue state tracking, \textbf{\textsc{MultiWOZ}2.1}~\cite{eric-etal-2020-multiwoz}.

Table~\ref{tab:statistics} shows the statistics of above datasets.
\subsection{Settings}

\begin{table*}[!t]
	\centering
	\small
	\begin{tabular}{lcccccccc}
		\toprule
		\multirow{3}{*}{\textbf{Model}}& \multicolumn{4}{c}{\textbf{data-v1}} &\multicolumn{4}{c}{\textbf{data-v2}} \\
		\cmidrule(lr){2-5} \cmidrule(lr){6-9}
		&\multicolumn{2}{c}{\textbf{dev}} &\multicolumn{2}{c}{\textbf{test}}
		&\multicolumn{2}{c}{\textbf{dev}} 
		&\multicolumn{2}{c}{\textbf{test}} \\
		&F1$(\delta)$ &F$1_c(\delta)$ &F1$(\delta)$ &F1$_c(\delta)$
		&F1$(\delta)$ &F$1_c(\delta)$ &F1$(\delta)$ &F1$_c(\delta)$\\
		\midrule
		GDPNet &67.1 (1.0) &61.5 (0.8) &64.9 (1.1) &60.1 (0.9) &- &- &- &- \\
		TUCORE-GCN &- &- &- &- & 66.8 (0.7) & 61.0 (0.5) & 65.5 (0.4) & 60.2 (0.6) \\
		TSP &66.8 (0.9) &61.5 (1.0) &65.5 (0.7) &60.5 (0.8) &- &- &- &- \\
		BERT &60.6 (1.2) &55.4 (0.9) &58.5 (2.0) &53.2 (1.6) &59.4 (0.7) &54.7 (0.8) &57.9 (1.0)	&53.1 (0.7)\\
		{BERT$_s$} &{63.0 (1.5)} &{57.3 (1.2)} &{61.2 (0.9)} &{55.4 (0.9)} &62.2 (1.3) &57.0 (1.0) &59.5 (2.1) &54.2 (1.4) \\
		BERT$_{c}$ &66.8 (0.9) &60.9 (1.0) &66.1 (1.1) &60.2 (0.8) &66.2 (0.9) &60.5 (1.1) &65.1 (0.8) &59.8 (1.2) \\
		\textsc{RoBERTa} &68.0 (1.0) &60.3 (1.0) &66.0 (0.6) &59.6 (0.7)  &67.6 (0.8) &61.0 (0.7) &65.8 (1.0) &59.6 (0.5) \\
		\midrule
		\rowcolor{mygray}
		\rowcolor{mygray}
		{SARA-BERT} &68.1 (1.0) &62.1 (0.9) &67.5 (0.7) &61.4 (0.9) &68.0 (0.8) & 62.1 (0.6) &67.3 (1.0) &61.3 (0.8) \\
		\rowcolor{mygray}
		{SARA}-\textsc{RoBERTa} &\textbf{69.3} (0.9) & \textbf{62.3} (0.8) &\textbf{68.1} (0.8) & \textbf{61.7} (1.0) &\textbf{69.5} (0.7) & \textbf{62.4} (0.5) &\textbf{67.8} (0.8) & \textbf{61.5} (0.7) \\
		\bottomrule
	\end{tabular}
	\caption{Performance on DialogRE. We report the average and the standard deviation computed from 5 runs, best results are marked in \textbf{bold}.}
	\label{tab:mainRE}
\end{table*}

\noindent\textbf{Model Configuration.} We take BERT-base and \textsc{RoBERTa}-base as our backbone model. For Pre-training, AdamW~\cite{loshchilov2018decoupled} is used as an optimizer, with an initial learning rate of $1 \times 10^{-5}$. We reduce the learning rate according to a linear scheduler.
The batch size is 2048, and the maximum input sequence length is 512. 
For the hyper-parameters, we set $\alpha=0.1, \beta=1.0, \tau=1.0$ in our experiments.
The pre-training of our model is carried out on 8 Nvidia Telsa V100 32G GPU for 5 epochs, taking about 2 days to reach convergence.
For fine-tuning, we follow previous works to set hyper-parameters.
More details can be found in Appendix~\ref{sec:appendix-c}.

\noindent\textbf{Metrics.} We use macro F1 and macro F1$_c$ for dialogue relation extraction (DialogRE), following~\citet{yu-2020-dialogue}. 
For intent prediction ({\textsc{banking77}}, \textsc{clinc150}, \textsc{hwu64}), we report the accuracy. 
Macro F1~\cite{coope-etal-2020-span} is adopted for slot filling tasks (\textsc{restaurant8k}, \textsc{DSTC8}).
For TOP, we use exact-match, which measures how often the model generates the exact reference structure.
For \textsc{MultiWOZ}, we use the joint goal accuracy following~\citet{budzianowski-etal-2018-multiwoz}.

\subsection{Compared Models}
For Dialogue relation extraction, we compare the proposed model with BERT-based models: \textbf{BERT} takes a pre-trained BERT as the dialogue encoder and predicts relation labels using the hidden state of the \texttt{[CLS]} token. \textbf{BERT$_s$}~\cite{yu-2020-dialogue} enhances the speaker representation by marking speaker arguments with special tokens. \textbf{BERT$_c$}~\cite{bai-etal-2021-semantic} concatenates hidden states of the \texttt{[CLS]} token and entity tokens for classification. 
For completeness, we also include recent methods which give the state-of-the-art results, such as \textbf{GDPNet}~\cite{xue20GDPNet}, \textbf{TUCORE-GCN}~\cite{lee-choi-2021-graph}, \textbf{TSP}~\cite{zhao-etal-2021-enhancing} and \textbf{Hier}~\cite{bai-etal-2021-semantic}.
We follow the implementation and hyper-parameters of {BERT$_c$} to evaluate our model.

For DialoGLUE, the compared models include: \textbf{BERT}~\cite{devlin-etal-2019-bert} pre-trains a Transformer encoder on large-scale monotonic text. \textbf{USE}~\cite{YangCAGLCAYTSSK20} pre-trains a dual Transformer encoder model on multilingual corpus using retrieval focused training tasks.
\textbf{\textsc{ConveRT} (654M)}~\cite{HendersonCMSWV20} pre-trains a dual Transformer encoder on the full 2015-2019 Reddit data comprising 654M $\left\langle\textit{context}, \textit{response}\right\rangle$ training pairs using response selection task.
\textbf{\textsc{ConvBERT} (700M)}~\cite{mehri2020dialoglue} fine-tunes a BERT model on 700M Reddit conversational data. 
We adopt the same implementation and hyper-parameters of {\textsc{ConvBERT} (700M)} to conduct experiments on DialoGLUE.

To verify the scalability of the proposed method, we also report results based on the \textsc{RoBERTa} model for all tasks.
The model architectures for about tasks is given in Appendix~\ref{sec:appendix-f}.

\begin{table*}[!t]
	\centering
	\small
	\begin{tabular}{lcccccccc}
		\toprule
		\textbf{Model} & \textbf{\textsc{bank}} &\textbf{\textsc{hwu64}} &\textbf{\textsc{clinc150}} &\textbf{\textsc{rest8k}} & \textbf{\textsc{DSTC8}} & \textbf{\textsc{TOP}} & \textbf{\textsc{MultiWOZ}}  &\textbf{Avg}\\
		\midrule
		USE & 92.81 &91.25 &95.06 & - &- &- &- &- \\
		\textsc{ConveRT} (654M) &93.01 &91.24  &97.16 & - & - & - & - &- \\
		USE+\textsc{ConveRT} (654M) &93.36 &\textbf{92.62} &\textbf{97.16} & - & - & - & -  &- \\
		\textsc{ConvBERT} (700M) & 93.44 &{92.38} &{97.11} & 95.44 & 91.20 & {82.08} & 56.56 &86.89\\
		BERT  &93.02 &89.87 &95.93 & 95.53 &90.05 & 81.90 & 56.30 &86.08 \\
		\textsc{RoBERTa} & 93.16 &91.30 & 96.09 & 96.27 & 90.78 & 81.80 &54.95  &86.28 \\
		\midrule
		\rowcolor{mygray}
		{SARA-BERT (6M)}  &{93.47} &92.01 &96.24 &{95.92} &{91.57} &82.05 &\textbf{59.33} &87.23 \\
		\rowcolor{mygray}
		{SARA-\textsc{RoBERTa} (6M)} &\textbf{93.64} &92.29 &96.60 &\textbf{96.74} & \textbf{92.02} &\textbf{82.78} & 57.52 &\textbf{87.37} \\
		\bottomrule
	\end{tabular}
	\caption{Performance on DialoGLUE, best results are in \textbf{bold}. \textsc{rest8k} and \textsc{bank} stands for \textsc{restaurant8k} and  \textsc{banking77}, respectively.}
	\label{tab:mainDGLUE}
\end{table*}

\begin{table}[!t]
	\centering
	\small
	\begin{tabular}{lccc}
		\toprule
		{\textbf{Model}} &{\textbf{DialogRE}} & {\textbf{\textsc{DSTC8}}}\\
		\midrule
		\textsc{RoBERTa} &67.6 &93.98  \\
		\textsc{RoBERTa} (6M) &68.2 &94.17 \\
		SARA-\textsc{RoBERTa} (6M)&\textbf{69.5} &\textbf{95.24} \\
		\quad w/o \text{sem$\_$mlm} &69.0 &95.01  \\
		\quad w/o \text{rel$\_$pred} &68.6 &94.63  \\
		\quad w/o \text{sem$\_$agree} &68.8 &94.72 \\
		\bottomrule
	\end{tabular}
	\caption{Validation F1 of DialogRE and \textsc{DSTC8}.}
	\label{tab:ablation}
\end{table}

\subsection{Main Results}
\noindent\textbf{Results on DialogRE.}
Table~\ref{tab:mainRE} lists the results of different systems on DialogRE.
Among BERT-based models (i.e., BERT, BERT$_s$, BERT$_c$), BERT$_c$ reports the best results.
Compared with BERT$_c$, SARA-BERT gives significantly ($p<0.001$) better results on both datasets.
In particular, SARA-BERT improves BERT$_c$ by 1.4 and 2.2 points in terms of F1 score on two test sets, respectively, indicating that our semantic pre-training framework is beneficial for dialogue relation extraction. 
The main reason can be that SARA improves the model capacity of understanding entities (which are core semantic units) and the semantic relations between them during pre-training stage.

SARA-BERT achieves better F1 scores than the other state-of-the-art methods.
In addition, when using \textsc{RoBERTa} as the backbone, SARA gives consistent improvements.
In particular, SARA-\textsc{RoBERTa} achieves 68.1 and 67.8 F1 scores on the test set of data-v1 and data-v2, respectively. 
To our best knowledge, these are the best-reported results based on \textsc{RoBERTa}-base.

\noindent\textbf{Results on DialoGLUE.}
We report the results of different methods on the DialoGLUE benchmark in Table~\ref{tab:mainDGLUE}.
Compared with BERT, SARA-BERT (6M) gives consistently better results on all $7$ datasets, with an improvement of $1.1$ point in average. 
In particular, SARA-BERT (6M) outperforms BERT by $2.1$ and $3.0$ points on \textsc{hwu64} and \textsc{MultiWOZ}, respectively, showing that our SARA framework can benefit task-oriented dialogue systems. 

Compared with the other state-of-the-art systems, SARA-BERT (6M) obtains better results than USE, because SARA-BERT (6M) is pre-trained on large-scale dialogue corpus.
In addition, SARA-BERT (6M) gives highly competitive results than \textsc{ConveRT} (654M), USE+\textsc{ConveRT} (654M) and \textsc{ConvBERT} (700M), using significantly fewer data (about $1\%$ than others). 
This indicates that our semantic-based pre-training framework is more data-efficient.
Finally, similar to SARA-BERT (6M), {SARA-\textsc{RoBERTa} (6M)} significantly ($p<0.001$) outperforms \textsc{RoBERTa}, giving the best results on \textsc{banking77}, \textsc{rest8K}, \textsc{DSTC8} and \textsc{TOP}.


\section{Analysis}
\subsection{Ablation Study}
We compare our full system with the following models: \textsc{RoBERTa} (6M) is continuously pre-trained on the exact same training corpus as our model using corresponding standard pre-training objectives; w/o \text{sem$\_$mlm}, w/o \text{rel$\_$pred}, and w/o \text{sem$\_$agree} denote the models which are trained without the semantics-guided masking, semantic relation prediction, and semantic agreement task, respectively.
Table~\ref{tab:ablation} shows the F1 scores on the validation sets of DialogRE and DSTC8.
First of all, using dialogue domain data (\textsc{RoBERTa} v.s. \textsc{RoBERTa} (6M)) for pre-training leads to improvements on both tasks. This meets previous observations~\cite{dontstoppretraining2020,mehri2020dialoglue}.
Also, the semantic-based mask language modeling task (\text{sem$\_$mlm}) gives an obvious improvement on DialogRE and a small one on DSTC8. 
The reason can be that DSTC8 has an average length of 8 tokens, making it easy to understand core semantic units in dialogues.
In addition, the performance drops significantly without the relation prediction task (rel$\_$pred), indicating that the rel$\_$pred task is important for dialogue understanding.
Furthermore, the semantic agreement task (sem$\_$agree) is helpful for both datasets, showing that the AMR is beneficial to improve the overall semantic representation of dialogue.
Finally, by combining dialogue domain data and all pre-training tasks, our final model achieves the best performance on both datasets.

\subsection{Effect of Semantic-based Pre-training}
\begin{figure}[!t]
	\setlength{\belowcaptionskip}{-0.1cm}
	\centering 
	\includegraphics[width=0.95\hsize]{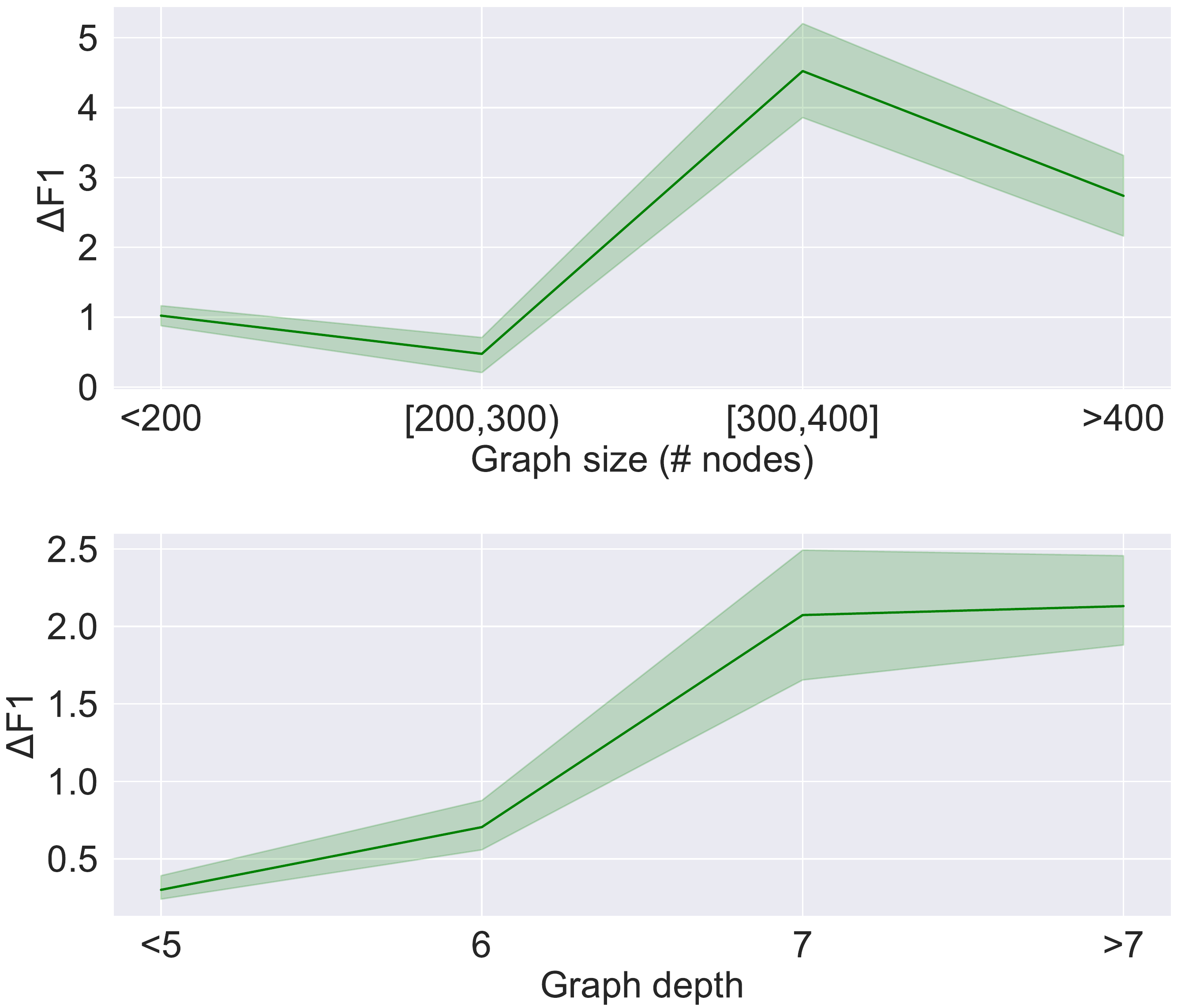}
	\caption{Performance improvement ($\Delta$F1) over two aspects: (top) graph size and (bottom) graph depth.}
	\label{fig:graph-feature}
\end{figure}
To further understand the effectiveness of our semantic-based pre-training framework, we split the test set of DialogRE (v2) into different groups according to semantic complexity and report the performance improvement of SARA-\textsc{RoBERTa} over \textsc{RoBERTa}.
In particular, two metrics are considered to measure the semantic complexity of a dialogue: 1) graph size (i.e., the number of nodes in the AMR graph) which records the number of semantic units in the dialogue; 2) graph depth which is defined as the longest distance between the AMR node and root node. 
An AMR graph has a deeper depth means that its corresponding dialogue has more long-range dependencies.

As shown in the top sub-figure of Figure~\ref{fig:graph-feature}, SARA-\textsc{RoBERTa} gives consistent improvements over \textsc{RoBERTa} in different graph groups. 
In particular, the improvements are more considerable on graphs with more than 300 nodes, showing that  SARA-\textsc{RoBERTa} has better capacity than \textsc{RoBERTa} in understanding dialogues which contain more semantic units.
The reason can be that the semantic-based MLM task enhances the model ability to capture core semantic features, which helps in reducing negative impacts of meaningless tokens in dialogue text.
With respect to graph depth, SARA-\textsc{RoBERTa} also outperforms \textsc{RoBERTa} on all groups, and larger improvements are observed on deeper graphs.
It can be that the relation prediction task helps to establish semantic associations between non-neighbor words, thus benefiting long-range dependencies understanding.

\begin{figure}[!t]
	\setlength{\belowcaptionskip}{-0.1cm}
	\centering 
	\includegraphics[width=0.85\hsize]{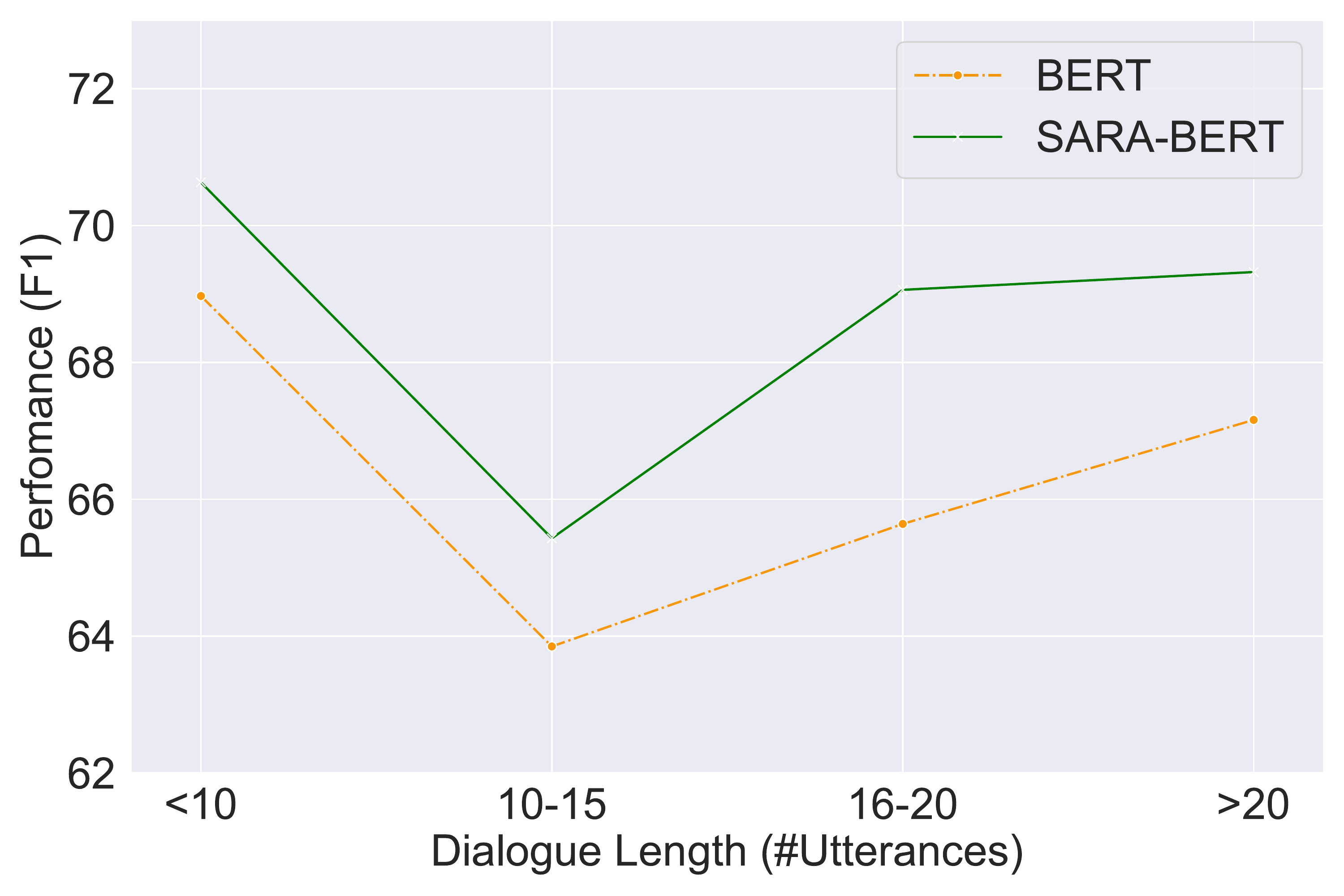}
	\caption{Test F1 on DialogRE (v2).}
	\label{fig:utterlen}
\end{figure}
We also compare the model performance in terms of dialogue length. 
In particular, we split the test set of DialogRE (v2) into $4$ groups according to the utterance number of each dialogue, and compare the performance of \textsc{RoBERTa} and SARA-\textsc{RoBERTa}.
As shown in Figure~\ref{fig:utterlen}, SARA-\textsc{RoBERTa} consistently gives better results than \textsc{RoBERTa} on all groups.
The performance gap is bigger when the input dialogue has more than 16 utterances. 
The reason is that SARA encourages the model to understand core semantics, which is helpful for learning long dialogues.

\subsection{Impact of AMR Features}
\begin{table}
	\centering
	\small
	\begin{tabular}{lcc}
		\toprule
        \textbf{Model} & DialogRE & DSTC8 \\
		\midrule 	
		\textsc{RoBERTa} &65.8 & 90.78  \\
		{SARA-\textsc{RoBERTa} (full)} &\textbf{67.8} &\textbf{92.02} \\
		{SARA-\textsc{RoBERTa} (simplified)} &67.3 &91.34 \\
		\bottomrule
	\end{tabular}
	\caption{F1 on the test set of DialogRE and \textsc{DSTC8}.}
	\label{tab:AMRvsShallow}
\end{table}
AMR is a deep semantic structure which consists of both backbone  relations and fine-grained semantic relations.
To study the contribution of such features, we simplify an AMR graph by masking the fine-grained semantic relations, resulting in a graph with frame arguments relations (e.g., \textit{:arg0, :arg1, :arg2}). 
We use the simplified graph as explicit semantic knowledge for pre-training and compare it with the standard AMR graph under the same framework.

Table~\ref{tab:AMRvsShallow} lists the performance of two systems. 
It can be observed that both \textit{simplified} graphs and \textit{full} AMR graphs lead to better performance.
Compared with \textit{simplified} graphs, using \textit{full} AMR graph for pre-training leads to better results on both DialogRE and DSTC8, showing that the fine-grained semantic features can further improve the model performance. 

\subsection{Comparison with explicit AMR}
\begin{figure}[!t]
	\setlength{\belowcaptionskip}{-0.1cm}
	\centering 
	\subfigure[]{\includegraphics[width=0.54\hsize]{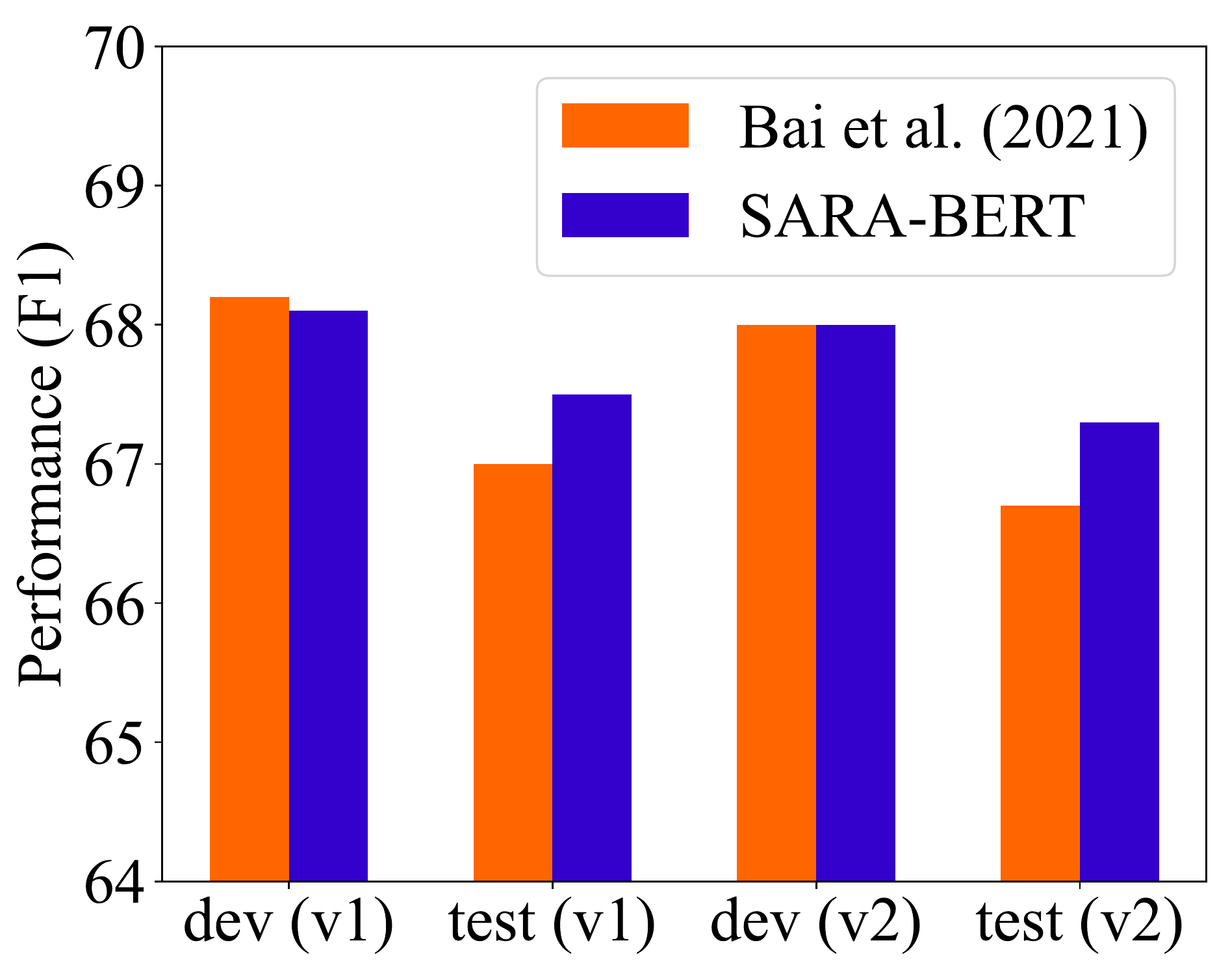}\label{fig:performance}}
	\subfigure[]{\includegraphics[width=0.42\hsize]{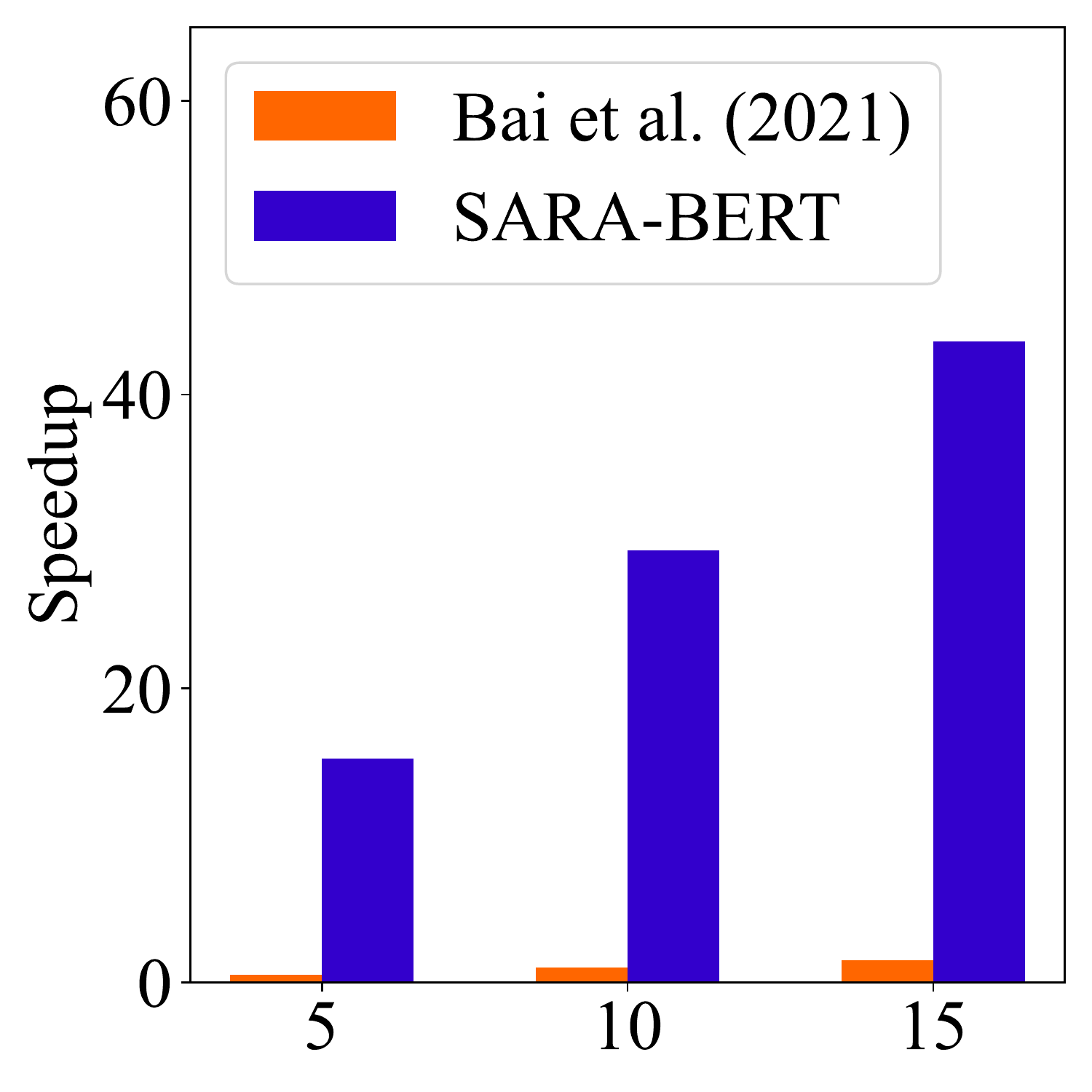}\label{fig:speed}} 
	\caption{(a) Comparison of performance on DialogRE; (b) Comparison of inference speed regarding to dialogue length (measured by number of utterances).}
	\label{fig:comparison}
\end{figure}

Figure~\ref{fig:performance} compares the performance of our model with the method of~\citet{bai-etal-2021-semantic}\footnote{We choose the \texttt{Hier} model which has comparable parameters to our model.} which use explicit AMR structures for dialogue applications. 
We report the F1 score on the test set of DialogRE.
Compared with the system of~\citet{bai-etal-2021-semantic}, our model gives comparable results on the validation set, and better results on the test set, without using an external AMR parser.
This indicates that 1) our pre-training framework can efficiently transfer the learned semantic information to downstream tasks; 2) large-scale semantic-aware pre-training can give further improvement compared with using semantic information in downstream tasks.

As shown in Figure~\ref{fig:speed}, our system is significantly faster than the method of~\citet{bai-etal-2021-semantic} which relies on an external parser.
As the dialogue length increases, the performance gap is more obvious.
In particular, our system obtains about a 45 times speedup when the input dialogues have an average utterance number of 15.

\subsection{Impact of Training Data Scale}
\begin{figure}[!t]
	\setlength{\belowcaptionskip}{-0.1cm}
	\centering 
	\includegraphics[width=0.9\hsize]{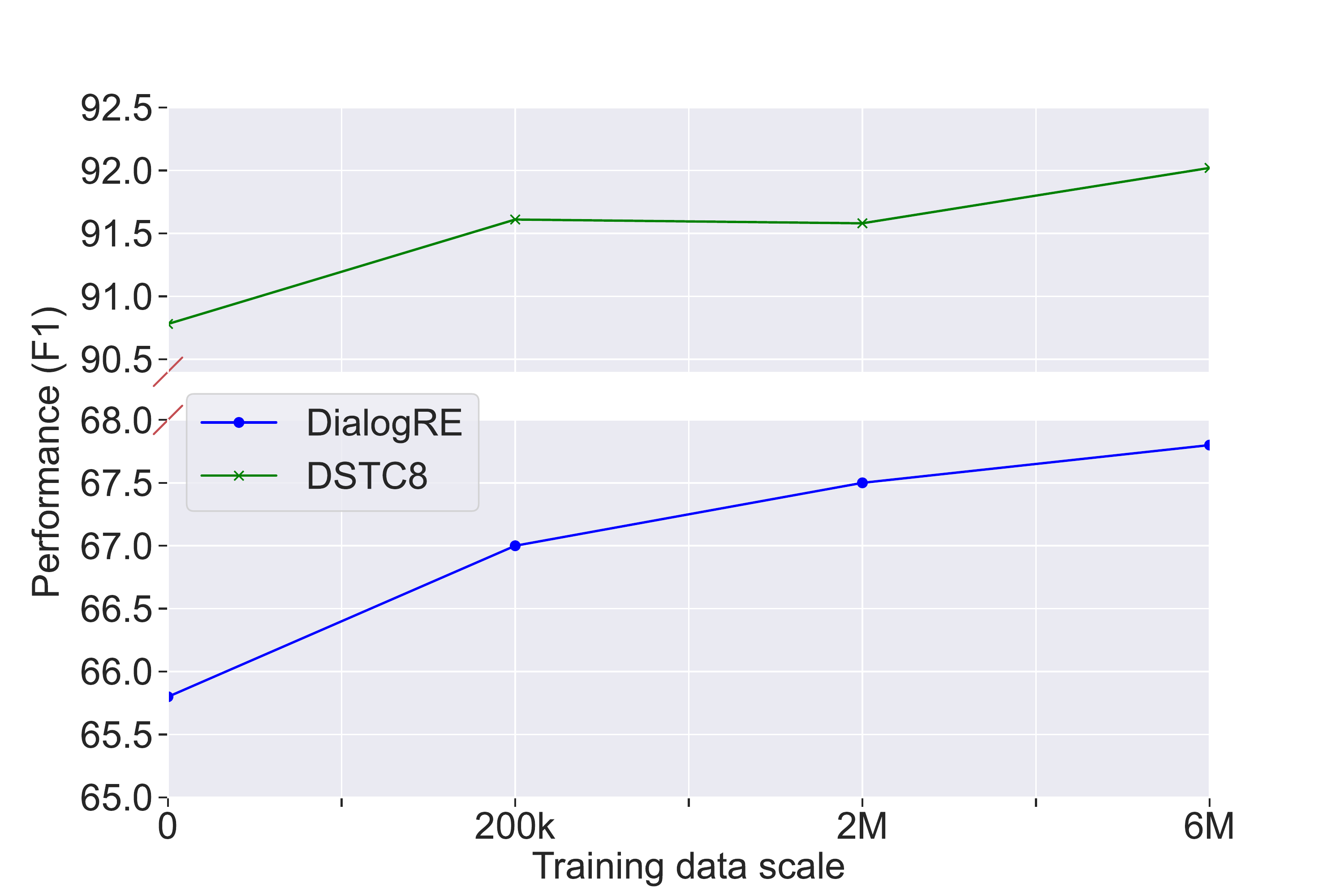}
	\caption{Impact of pre-training data.}
	\label{fig:data-scale}
\end{figure}
Figure~\ref{fig:data-scale} shows the model performance regarding different scales of pre-training data.  
The performance on both DialogRE and DSTC8 datasets increases as the
scale of training data grows bigger, with a margin of about 2.0 F1 score on DialogRE.
Due to the limitation of computational resources, we do not conduct experiments on larger training corpus and models, and we leave this for a future work.

\subsection{Case Study}
\label{sec:casestudy}

\begin{figure}[!t]
	\setlength{\belowcaptionskip}{-0.1cm}
	\centering 
	\includegraphics[width=0.9\hsize]{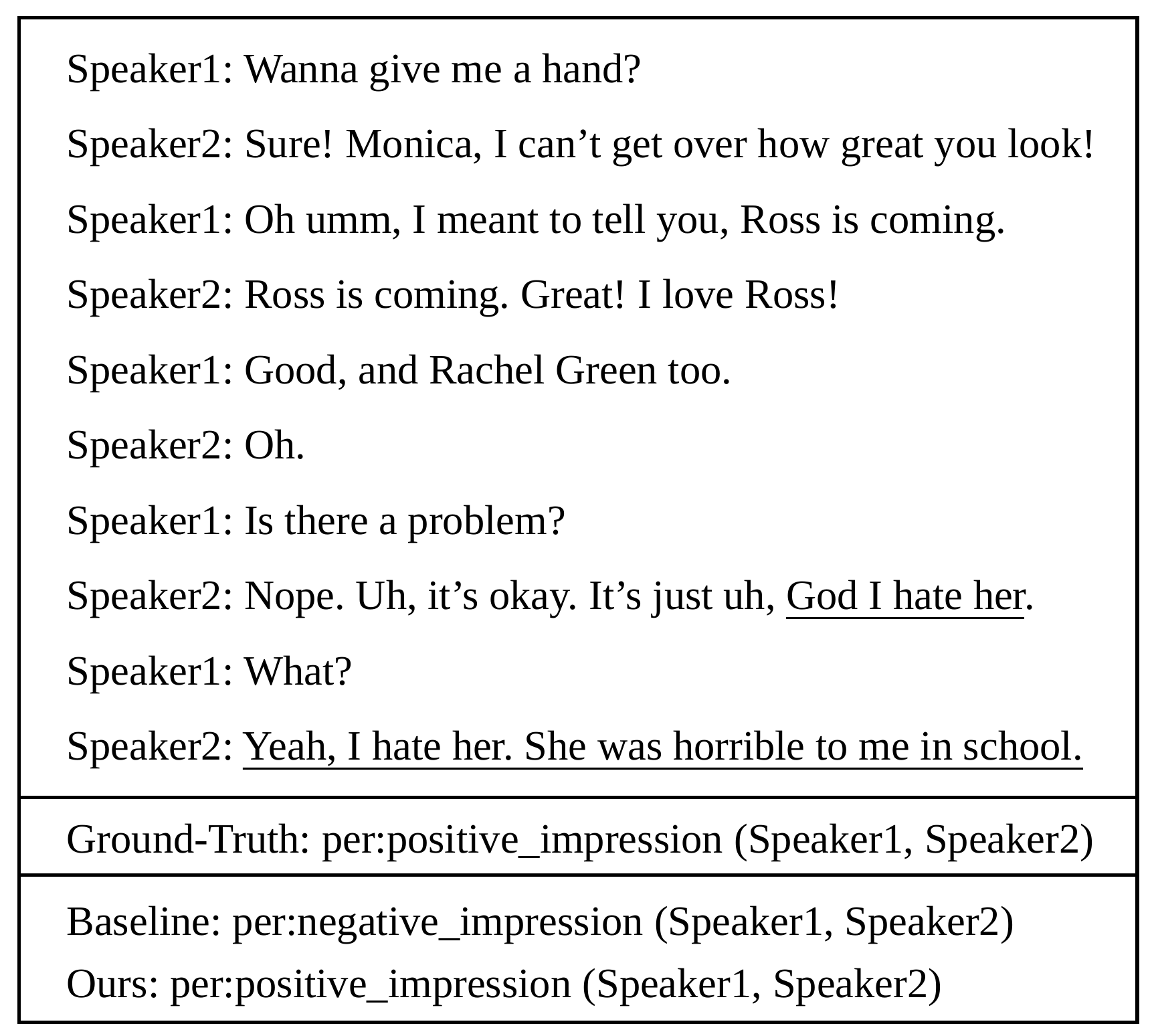}
	\caption{An example of dialogue relation extraction.}
	\label{fig:case}
\end{figure}

Figure~\ref{fig:case} shows an example conversation from DialogRE dataset.
The baseline model (\textsc{RoBERTa}) is misled by sentences last three utterances (marked with underline) where \textit{Speaker2} shows an negative emotions towards Rachel Green, and thus incorrectly predicting the relationship between two speakers as \textit{negative$\_$impression}.
In contrast, our model (SARA-\textsc{RoBERTa}) predicts the correct relationship, suggesting that our semantic-based pre-training framework helps model to better understand the relationship between entity pairs and avoid focusing on spurious features. 

\begin{figure}
	\setlength{\belowcaptionskip}{-0.1cm}
	\centering 
	\includegraphics[width=0.9\hsize]{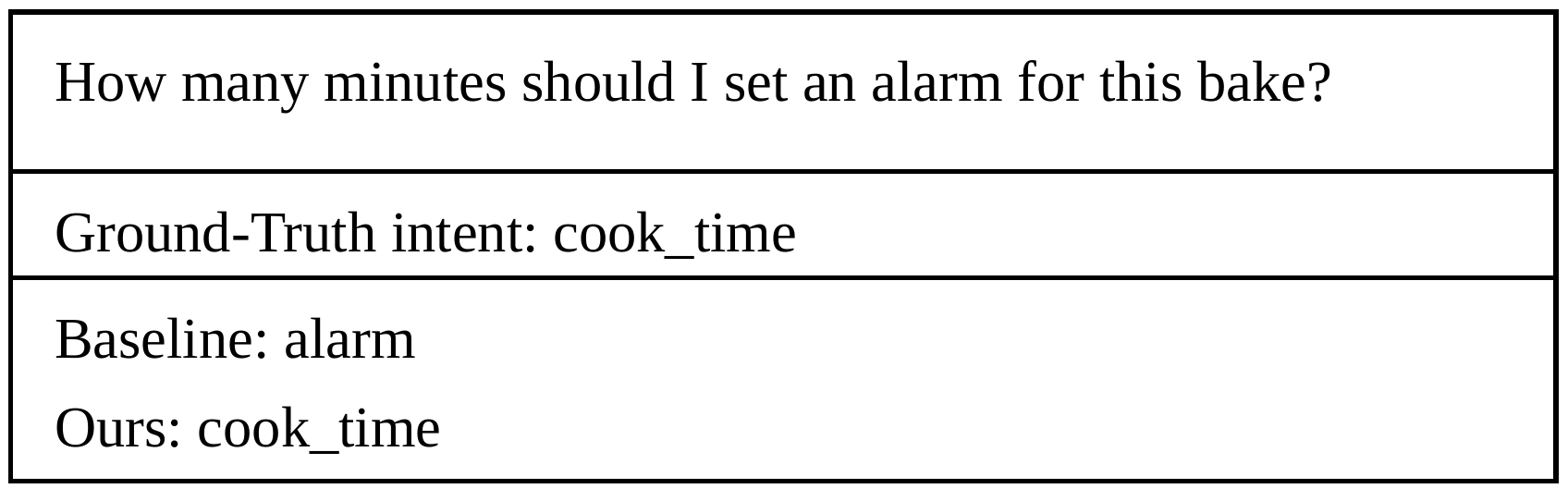}\label{fig:case3}
	\caption{An example of dialogue intent prediction.}
	\label{fig:case-intent}
\end{figure}

Figure~\ref{fig:case-intent} presents a case of dialogue intent prediction. 
The baseline system pays much attention on word ``\textit{alarm}'' while ignores other two core semantic units ``\textit{minutes}'' and ``\textit{bake}'', giving an incorrect prediction.
Our system successfully predicts the gold intent, because AMR guides our model to discover the core semantic units in the dialogue text.

\section{Conclusion}
We investigated the abstract meaning representation as explicit semantic clues for dialogue pre-training, using a semantic-based pre-training framework.
Experiments on two benchmarks show that the proposed framework is highly effective on both chit-chat understanding and task-oriented dialogue understanding.
Our method gives the best results on multiple datasets.

\section*{Acknowledgments}
Yue Zhang is the corresponding author. 
We would like to thank anonymous reviewers for their insightful comments.
This work is supported by the Zhejiang Province Key Project 2022SDXHDX0003 and the Tencent AI Lab Rhino-Bird Focused Research Program.

\bibliography{custom}
\bibliographystyle{acl_natbib}

\clearpage
\appendix

\section*{Appendix}

\section{Data Pre-processing}
\label{sec:appendix-a}
For pre-training, we randomly sample 10 million dialogue from Reddit~\cite{henderson-etal-2019-repository} corpus and filter the data by removing the instances where
\begin{itemize}
    \item dialogue contains special markers;
    \item dialogue contains more than 10 non-English tokens;
    \item dialogue is longer than 150 words;
    \item dialogue has more than 15 turns.
\end{itemize}
We also replace the URLs in dialogues with a special token \textit{<url>}.

\section{Model Input Format}
\label{sec:appendix-b}
Take BERT-based model as an example, given a dialogue $\bm{x}$ which consists of $n$ utterances, we concatenate all utterances as a single consecutive token sequence with special tokens separating them: $\bm{x}=$ \big{\{}[\texttt{CLS}] [\texttt{Utter}$_1$] \texttt{Speaker}$_1$ $U_1$ [\texttt{Utter}$_2$] \texttt{Speaker}$_2$ $U_2$ \dots [\texttt{Utter}$_n$] \texttt{Speaker}$_n$ $U_n$[\texttt{SEP}]\big{\}}, where $U_1$, $U_2$, $U_n$ are utterance sequences. [\texttt{CLS}] and [\texttt{SEP}] mark the start and end of the dialogue. [\texttt{Utter}$_1$], [\texttt{Utter}$_2$], and [\texttt{Utter}$_n$] mark the utterance numbers. \texttt{Speaker}$_1$ denotes the speaker of the first utterance. For \textsc{RoBERTa}, we use \texttt{<s>} and \texttt{</s>} to surround the dialogue sequences.

\section{Model Hyper-Parameters}

\label{sec:appendix-c}

\begin{table}[!t]
    \centering
    \small
    \begin{tabular}{l|c}
        \toprule
        \textbf{Param. Name} & \textbf{Value} \\
        \midrule
        Batch Size &2048 \\
        Optimizer & AdamW \\
        Learning Rate (lr) & 1e-5 \\
        Lr Scheduler & linear  \\
        Warmup Step & 0 \\
        Max Training Epoch & 5 \\
        Semantic Masking Prob. & 0.2 \\
        Extended Vocabulary Size & 30,774 \\
        Max Length (dialogue) & 256 \\
        Max Length (AMR) & 512 \\
        Mix Precision & fp16 \\
        Parameters (Pre-training) &219M \\
        Parameters (downstream tasks) &110M \\
        Training Time & about 45h \\
        \bottomrule
    \end{tabular}
    \caption{Hyper-parameters of our models.}
    \label{tab:hyperparams}
\end{table}

Table~\ref{tab:hyperparams} lists all model hyper-parameters used
for our experiments. The proposed model is implemented
based on \textit{Pytorch} and \textit{Huggingface Transformers}\footnote{https://github.com/huggingface/transformers}.
Our source code and pre-trained models is released at \url{https://github.com/goodbai-nlp/Sem-PLM}.

\section{Architecture for Downstream Tasks}
\label{sec:appendix-f}
For all downstream dialogue understanding tasks, we use the pre-trained dialogue model as a dialogue encoder and make prediction based on the encoded hidden states. 
Taking the BERT-based model as an example, the model architecture of downstream task are:

\noindent\textbf{Dialogue Relation Extraction}: We concatenate the hidden states  of two entities (denoted as $e_1$ and $e_2$) as well as the pooled representation of the \texttt{[CLS]} token into a linear classifier to predict the relation label as:
\begin{equation}
    y = \texttt{MLP}_c([\texttt{pool}(h^{\texttt{[CLS]}}); vec(e_1); vec(e_2)]),
\end{equation}
where \texttt{MLP}$_c$ is a linear classifier, and \textit{vec}($\cdot$) selects the encoded representation of the input token. 
\texttt{pool}($h^{\texttt{[CLS]}}$) passes the hidden state of the \texttt{[CLS]} token through a linear layer.

\noindent\textbf{Intent Prediction}: We solve the task as a sequence classification problem, by feeding the pooled hidden state of \texttt{[CLS]} token into a linear classifier to predict the relation label as:
\begin{equation}
    y = \texttt{MLP}_c(\texttt{pool}(h^{\texttt{[CLS]}})).
\end{equation}

\noindent\textbf{Slot Filling}: We represent the
problem as IOB tagging, by feeding all hidden state of the input dialogue (denoted by $H$) into a linear classifier and predict the relation label as:

\begin{equation}
    Y = \texttt{MLP}_c(H),
\end{equation}
where $H$ denotes the output hidden states, and $Y$ is the output tag sequence.

\noindent\textbf{Semantic Parsing}: We solve the
problem as joint sequence classification and sequence labeling task.  
Specifically, we predict the intent and slots label as:
\begin{equation}
    \begin{split}
        y_{intent} &= \texttt{MLP}_{intent}(\texttt{pool}(h^{\texttt{[CLS]}})), \\
        Y_{slot} &= \texttt{MLP}_{slot}(H), \\
    \end{split}
\end{equation}
where $H$ denotes the output hidden states, and $Y$ is the output tag sequence.

\noindent\textbf{Dialogue State Tracking}: We follow the TripPy~\cite{HeckNLGLMG20} framework make prediction, which uses BERT model as encoder and combines BERT with a triple copy strategy to perform state tracking. Please refer the original paper for more details.
\end{document}